\documentclass[lettersize,journal]{IEEEtran}
\usepackage[utf8]{inputenc}
\usepackage{amsmath,amsfonts}
\usepackage{hyperref}
\usepackage{algorithmic}
\usepackage{algorithm}
\usepackage{bbding} 
\usepackage{color}
\usepackage{multirow}
\usepackage{array}
\usepackage[caption=false,font=normalsize,labelfont=sf,textfont=sf]{subfig}
\usepackage{textcomp}
\usepackage{stfloats}
\usepackage{url}
\usepackage{verbatim}
\usepackage{graphicx}
\usepackage{cite}
\usepackage{tablefootnote}

\usepackage{mathrsfs}

\begin{document}

\title{3DGTN: 3D Dual-Attention GLocal Transformer Network for Point Cloud Classification and Segmentation}

\author{
Dening Lu,
Kyle (Yilin) Gao,~\IEEEmembership{Graduate Student Member, IEEE,}
Qian Xie,
Linlin Xu,~\IEEEmembership{Member,~IEEE,}
Jonathan Li,~\IEEEmembership{Fellow,~IEEE}

\thanks{\textit{(Corresponding authors: Linlin Xu; Jonathan Li.)}}

\thanks{Dening Lu, Kyle Gao, Linlin Xu, and Jonathan Li are with the Department of Systems Design Engineering, University of Waterloo, Waterloo, Ontraio N2L 3G1, Canada (e-mail: {d62lu, y56gao, l44xu, junli}@uwaterloo.ca).}
\thanks{Jonathan Li is also with the Departmemt of Geography and Environmental Management, University of Waterloo, Waterloo, Ontario N2L 3G1, Canada.}
\thanks{Qian Xie is with the Department of Computer Science, University of Oxford, Oxford OX1 3QD, U.K. (e-mail: qian.xie@cs.ox.ac.uk).}
}

\markboth{}%
{Shell \MakeLowercase{\textit{et al.}}: Bare Demo of IEEEtran.cls for IEEE Journals}

\maketitle

\begin{abstract}
Although the application of Transformers in 3D point cloud processing has achieved significant progress and success, it is still challenging for existing 3D Transformer methods to efficiently and accurately learn both valuable global features and valuable local features for improved applications. This paper presents a novel point cloud representational learning network, called 3D Dual Self-attention Global Local (GLocal) Transformer Network (3DGTN), for improved feature learning in both classification and segmentation tasks, with the following key contributions. First, a GLocal Feature Learning (GFL) block with the dual self-attention mechanism (i.e., a novel Point-Patch Self-Attention, called PPSA, and a channel-wise self-attention) is designed to efficiently learn the GLocal context information. Second, the GFL block is integrated with a multi-scale Graph Convolution-based Local Feature Aggregation (LFA) block, leading to a Global-Local (GLocal) information extraction module that can efficiently capture critical information. Third, a series of GLocal modules are used to construct a new hierarchical encoder-decoder structure to enable the learning of "GLocal" information in different scales in a hierarchical manner. The proposed framework is evaluated on both classification and segmentation datasets, demonstrating that the proposed method is capable of outperforming many state-of-the-art methods on both classification and segmentation tasks. \textit{Our code will be made publicly available}.
\end{abstract}

\begin{IEEEkeywords}
Transformer, Graph convolution, Point cloud classification, Point cloud segmentation, Deep learning, Self-attention mechanism.
\end{IEEEkeywords}

\section{Introduction}
\label{sec:introduction}
\IEEEPARstart{P}{oint} cloud classification and segmentation are the fundamental tasks in 3D computer vision. Point clouds, being flexible, simple, and with easy-to-use data structures, are commonly used in 3D mapping, robotics, autonomous navigation, and city information modeling. With the ever-increasing availability and affordability of 3D sensors in the form of LiDAR scanners and RGB-D cameras, point cloud classification and segmentation thereof are ever more important. From the perspective of point cloud processing, both local features and global features play an important role in classification and segmentation tasks. Local features refer to the features that capture the local geometric patterns and details of the point cloud. Global features refer to the features that capture the overall shape and structure of the entire point cloud. A combination of global and local features (called \textit{GLocal} features here) is able to provide the model with a more complete representation of the target point cloud. 

For classification and segmentation tasks, many types of deep learning architectures were experimented with in the recent past. Among these, the Transformer \cite{2017transformer} architecture emerged as a powerful point cloud feature extraction backbone, performing exceedingly well on point cloud classification, detection, and segmentation~\cite{lu2022transformers}. 
First developed for natural language processing, the Transformer is a low-inductive bias network that is capable of learning long-range features. 
Since then, Transformers have successfully been applied to 2D and 3D computer vision to a wide variety of tasks, achieving state-of-the-art results across a wide variety of benchmarks. 

Specifically, some existing 3D Transformer methods like Point Transformer \cite{zhao2021point} proposed to utilize Transformer blocks for local feature aggregation in a hierarchical way. They have demonstrated the feasibility and effectiveness of the Transformer in the field of 3D point cloud processing. Other 3D transformer-based approaches, such as Point Cloud Transformer \cite{guo2021pct}, PatchFromer \cite{zhang2022patchformer}, and Point Attention Transformers \cite{yang2019modeling}, proposed well-designed self-attention variants for algorithm performance or efficiency improvement.
Although different 3D Transformer approaches demonstrated strong feature learning capabilities in 3D point cloud applications, they still have limitations in terms of modeling both the local information and global information in an efficient, accurate, and fast manner. 

This paper presents a 3D Dual-attention Global-Local (GLocal) Transformer Network, called 3DGTN, which is tailor-designed to improve GLocal feature learning in 3D point cloud data processing, with the following key characteristics. 

\begin{itemize}
    \item A GLocal Feature Learning (GFL) block with the dual self-attention mechanism (comprised of a novel Point-Patch Self-Attention, called PPSA, and a channel-wise self-attention) is designed to efficiently learn the GLocal context information. The PPSA approach can better capture global correlation among local neighborhoods. The dual-attention mechanism integrates PPSA and Channel-wise Self-Attention (CSA) to improve the learning of critical information in both the spatial domain and feature domain.
    \item The GFL block is integrated with a Local Feature Aggregation (LFA) block into a Global-Local (GLocal) information extraction module to enable the learning of both valuable global information and critical local information. The LFA block is designed based on the Graph Convolution Network (GCN) to improve both the efficiency and accuracy of local information extraction. 
    \item The GLocal modules are used to construct a new hierarchical encoder-decoder structure to enable the learning of ``GLocal" information at different scales in a hierarchical manner, leading to a general point cloud representation network that can improve both classification and segmentation tasks. Based on this framework, a 3D classification and a 3D semantic segmentation algorithm are designed and used to improve the point cloud classification and segmentation tasks.
    
\end{itemize}

Extensive experiments comparing the proposed approaches with many state-of-the-art algorithms on many datasets, i.e., ModelNet40, ShapeNet, and Toronto-3D demonstrate that our method exceeds previous state-of-the-art performance in both classification and segmentation tasks.


\section{Related Work}
\label{sec:relatedwork}

The application of transformers in 3D point cloud processing has yielded remarkable achievements, contributing to advancements in various fundamental tasks including classification, segmentation, and detection. 
Transformer-based methods tailored for point cloud data can be broadly categorized into two main groups: global Transformer-based methods and local Transformer-based methods.
Here, we review existing approaches in both categories and summarize the limitations. 


\subsection{Global Transformers in 3D Point Cloud Processing}
 
The global Transformer approaches focus on learning large-scale context information from the 3D point cloud to improve classification and segmentation. Point Cloud Transformer (PCT), as a standard global Transformer network, was proposed in \cite{guo2021pct}. In PCT, all input points were leveraged for global feature extraction. PCT first adopted a neighborhood-embedding strategy to aggregate the local information, followed by feeding the embedded features into four stacked global Transformer blocks. Lastly, it utilized a global Max and Average (MA) pooling to extract the global information for point cloud classification. The segmentation network of PCT \cite{guo2021pct} took its classification network variant as the encoder. The decoder first concatenated the pooled global feature with each point feature, enhancing the perception of global information for each point. Then the concatenated features were fed into a series of MLP layers for dense prediction, following PointNet \cite{qi2017pointnet}. 

3CROSSNet proposed in \cite{han20223crossnet} used multi-scale global information for classification. Taking the raw point cloud as input, it first generated three point subsets with different resolutions by Farthest Point Sampling (FPS). Secondly, it established $k$-Nearest Neighborhood ($k$NN) \cite{qi2017pointnet} and extracted local information by a series of Multi-Layer Perception (MLP) modules for each point subset. Thirdly, the cascaded global Transformer blocks were applied to extract the global information of each subset. Lastly, given the multi-scale global features, 3CROSSNet used the Cross-Level Cross-Attention (CLCA) and Cross-Scale Cross-Attention (CSCA) modules to capture long-range inter- and intra-level dependencies for classification. 

Instead of using raw point clouds, Stratified Transformer \cite{lai2022stratified} took 3D voxels as input to the segmentation network. It applied Transformers in predefined local windows, following Swin Transformer \cite{liu2021swin}. To capture the global information and establish connections between different windows, it presented a novel $key$ sampling strategy, enlarging the effective receptive field for each $query$ point. 

\subsection{Local Transformers in 3D Point Cloud Processing}
As a local Transformer network, Point Transformer (PT) \cite{zhao2021point} focused on extracting local information by the Transformer. It had five local Transformer blocks operated on the downsampling point sets. Specifically, for each block, it constructed $k$NN for sampling points, then utilized a vector-attention mechanism to capture local features. After five local Transformer blocks, PT used a global MA pooling to extract the global feature for classification. Local Feature Transformer Network (LFT-Net) \cite{gao2022lft} had a similar architecture. However, it used an additional trans-pooling module to alleviate the feature loss during the pooling. For 3D point cloud segmentation, PT \cite{zhao2021point} developed the segmentation network based on its classification framework. The authors designed a U-net style architecture for segmentation, where the decoder was symmetric to the encoder. Since it used a hierarchical structure in the encoder, a transition-up module with trilinear interpolation was proposed in the decoder for point cloud upsampling. 

\subsection{Limitations of Current 3D Transformer-based Networks}

Despite the great success of Transformers in point cloud classification and segmentation, existing 3D Transformer methods tend to only consider local information extraction or are still challenging to learn both global and local features effectively. For example, PT \cite{zhao2021point} only utilized Transformer blocks in local neighborhoods, while ignoring global feature learning. PCT \cite{guo2021pct} only captured local information at the beginning of the network, as data preprocessing. It cannot dynamically fuse the local information with the global information extracted from each stage in the network. Recently, there are several works \cite{lu20223dctn, feng2020point, hui2021pyramid} that extract both local and global features in a simple cascading way. However, it is easy for them to lose local neighborhood information, which weakens the learning of global and local information. Therefore, this paper proposes a novel PPSA (Sec. \ref{subsec:GFL}) mechanism to improve global and local feature learning.  It not only captures global features by self-attention but also fuses the global features and local neighborhood information effectively. Considering that the dual-attention approaches \cite{qiu2021geometric, han2022dual} have demonstrated stronger feature learning capability, we design novel GFL blocks by the dual-attention Transformer, which integrates PPSA and CSA mechanisms. The cooperation of LFA and GFL blocks allows our network to have a strong global and local feature learning ability to improve 3D point cloud classification and segmentation. 


\begin{figure*}[htbp]
  \centering
  \includegraphics[width=\linewidth]{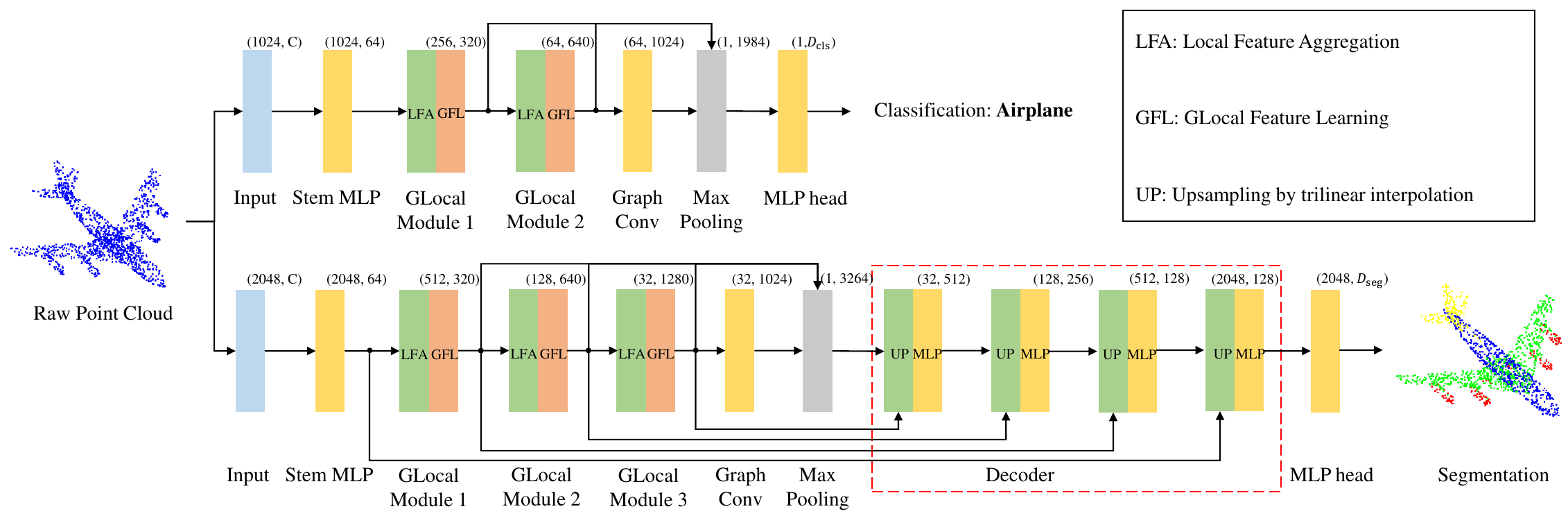}
  \caption{3DGTN networks for point cloud classification (top) and segmentation (bottom), where GCN-based LFA blocks and dual self-attention-based GFL blocks are designed for strong feature representation.
  \label{fig:overview}}
\end{figure*}

\section{3D Dual-attention GLocal Transformer}
\label{sec:method}
In this section, we introduce the encoder-decoder structure of our 3DGTN for both point cloud classification and segmentation. We first show the pipeline of our method, then introduce the main blocks in the encoder and decoder respectively.

\subsection{Overview}
Fig. \ref{fig:overview} shows the overall pipeline of our method. We propose a general backbone network for point cloud classification and segmentation.
Our classification and segmentation networks both share the same encoder architecture. 
After that, the classification network utilizes an MLP head to obtain the final classification results, while the segmentation network utilizes a decoder with trilinear interpolation-based upsampling for dense prediction.

The original point cloud with normal is taken as input to the encoder. We first design a stem MLP block to project the input data into a higher-dimension space. 
After that, the projected features are fed into stacked LFA and GFL blocks in a hierarchical manner for GLocal feature extraction. Specifically, the LFA block is achieved by the multi-scale GCN \cite{wang2019dynamic}, and the GFL block is achieved by the Transformer.
Following this, we use the max-pooling operation on the output feature maps of each module, to obtain the GLocal feature of each level. 
Then, we concatenate them for multi-level GLocal feature generation.
Given the extracted GLocal feature, we leverage an MLP head for the point cloud classification task, which consists of two fully connected layers with batch normalization and RELU activation. For the segmentation task, the extracted GLocal features are then taken as input to the decoder. 
To improve efficiency, we adopt the ALL-MLP decoder structure, instead of a symmetric one. In the upsampling block, the interpolated points are concatenated with the corresponding feature points from the encoder via a skip connection. We note that the number of modules in the encoder and decoder can vary according to the number of input points. In our experiments, we designed a two-module encoder for the classification task (1024 points), but a three-module encoder and corresponding decoder for the segmentation task (2048 points).

\begin{figure*}[htbp]
  \centering
  \includegraphics[width=0.9\linewidth]{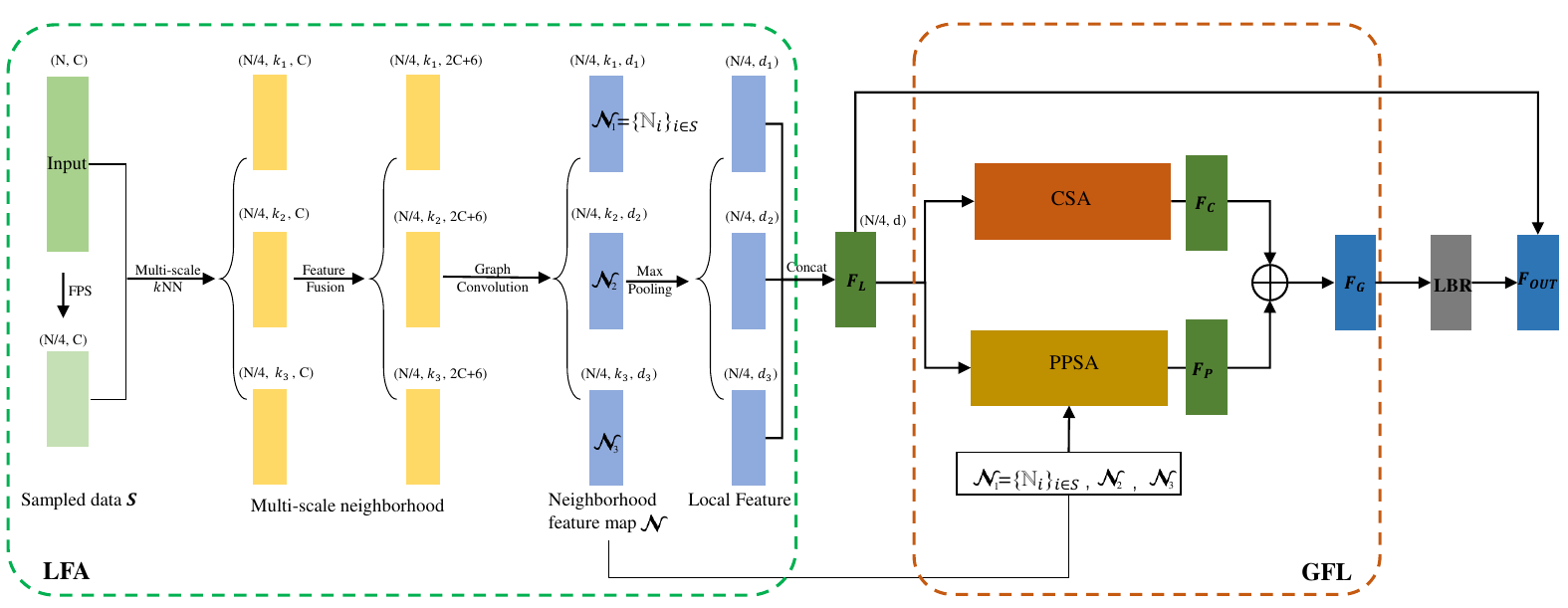}
  \caption{Architecture of GLocal Module 1, which consists of an LFA block and a GFL block.
  \label{fig:LFA}}
\end{figure*}

\subsection{Local Feature Aggregation Block}
\label{subsec:LFA}
We adopt the GCN-based LFA block for local feature aggregation. The LFA block (Fig. \ref{fig:LFA}) is introduced as follows.

The input point cloud is first downsampled to $N/4$ points by FPS, generating a sampled point subset $S$, where $N$ is the number of the input points. After that, the LFA block constructs multi-scale $k$-NN neighborhoods (three scales $k_{1}$, $k_{2}$, $k_{3}$ in our experiments) for each sampled point, to ensure the diversity of the receptive fields. In each neighborhood $\chi_{i}$ of the sampled point $S_{i}$, a fused feature $\mathbb{C}_{i}$ is generated by combining its coordinate and feature information $P_{i}, F_{i}$:

\begin{equation}
\mathbb{C}_{i}= concat (F_{i}, P_{i}),
\end{equation}
where $F_{i}$ and $P_{i}$ represent the semantic and geometric properties of $S_{i}$ respectively. 
Given the fused neighborhood feature, the Graph Convolution in $\chi_{i}$ can be formulated as:
\begin{equation}
l_{i} = \operatorname*{\textit{maxpooling}}\limits_{j \in \chi_{i}} \, (Conv(\Delta\mathbb{C}_{ij})),
\label{eq:1}
\end{equation}
where $l_{i}$ is the aggregated local feature of $S_{i}$, $Conv$ is a convolution operation with $1\times1$ kernels, $\Delta\mathbb{C}_{ij}$ represents the relationship between the $j$-th neighborhood point $\chi_{ij}$ and $S_{i}$, which is defined as:
\begin{equation}
\Delta \mathbb{C}_{ij}= concat(F_{ij} - F_{i}, P_{ij} - P_{i}, \mathbb{C}_{i}),
\end{equation}
where $F_{ij}$ and $P_{ij}$ represent the semantic and geometric properties of $\chi_{ij}$ respectively. Specifically, in Fig. \ref{fig:LFA}, since the dimensions of the input feature map and corresponding point set are $(N, C)$ and $(N, 3)$, the dimension of $\Delta \mathbb{C}_{ij}$ is $2C+6$. Furthermore, we define different output dimensions of Graph Convolution for different scale neighborhoods: $d_{1}$, $d_{2}$, and $d_{3}$, where $d_{1} < d_{2} < d_{3}$ ($k_{1} < k_{2} < k_{3}$).
$Conv(\Delta\mathbb{C}_{ij})$ establishes semantic and geometric relationships between the sampling point $S_{i}$ and neighborhood point $\chi_{ij}$. As such, a neighborhood feature set containing local information, $\mathbb{N}_{i}$ of $S_{i}$, is generated. Then, the max-pooling operation is used to aggregate the local information to $S_{i}$.
Given the aggregated local feature $l_{i}$ of $S_{i}$ at each scale, we finally concatenate them to generate the multi-scale local feature $L_{i}$ of $S_{i}$, which can be expressed as:

\begin{equation}
L_{i} = concat(l_{i1}, l_{i2}, l_{i3}),
\end{equation}
where $l_{i1}, l_{i2}, l_{i3}$ denote three local features of $S_{i}$ at three different scales.

\begin{figure*}[htbp]
  \centering
  \includegraphics[width=0.9\linewidth]{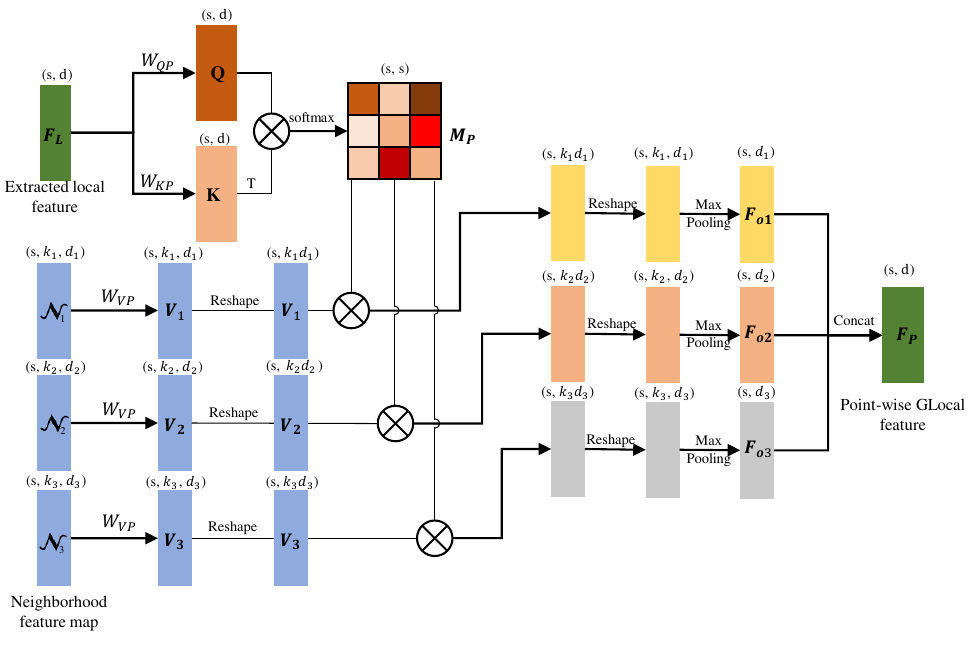}
  \caption{Point-Patch Self-Attention mechanism. It utilizes the sampling point features and the corresponding neighborhood (patch) feature maps for point-wise GLocal feature extraction.
  \label{fig:PSA}}
\end{figure*}

\subsection{GLocal Feature Learning Block}
\label{subsec:GFL}
Our GFL block contains two kinds of self-attention mechanisms: PPSA and CSA. PPSA, as a novel point-wise self-attention mechanism, is proposed to fuse the global features and local neighborhood information extracted from the LFA block for better GLocal feature learning. CSA is utilized to measure the correlation among different feature channels. It is able to improve context information modeling by highlighting the role of interaction across various channels. A detailed introduction to these two mechanisms is as follows.

\textbf{Point-Patch Self-Attention.} PPSA, as a novel point-wise self-attention mechanism, is designed to not only utilize the sampling points but also leverage their neighborhood (patch) information for GLocal feature learning.
As shown in Fig. \ref{fig:PSA}, the aggregated features
$F_{L}=\left\{ L_{i} \right\}_{i \in s} \in R^{s \times d}$ from the LFA block is taken as input, where $s$ is the number of sampled points in $S$, and $d$ denotes the feature dimension of $F_{L}$. We first project $F_{L}$ into two different feature spaces to generate $Query$, $Key$ matrices:
\begin{equation}
\begin{aligned}
Query =  F_{L} W_{QP},\\
Key =  F_{L} W_{KP} ,\\
\end{aligned}
\end{equation}
where $W_{QP}, W_{KP}$ are learnable weight matrices. Then, the attention map $M_{P}\in R^{s \times s}$ of PPSA can be formulated as:
\begin{equation}
M_{P} = softmax(\frac{QK^{T}}{\sqrt{d}}+B),
\end{equation}
where $Q,K$ denote the $Query$, $Key$ matrices, and $B$ is a learnable position encoding matrix defined by \cite{zhao2021point}. 
Next, we treat the neighborhood feature map $\mathcal{N} = \left\{ \mathbb{N}_{i} \right\}_{i \in s}$ at each scale as the $Value$ branch, instead of $F_{L}$ used by the vanilla PSA. In other words, the elements in the attention map are taken as weights of the corresponding neighborhood feature sets in $\mathcal{N}$. Then, the output neighborhood feature set is obtained by computing a weighted sum of all input sets. 
As such, we leverage all the points including sampling points and neighborhood points for the GLocal information extraction, instead of only sampling points. This method is able to improve the GLocal feature learning ability and mitigate the local information loss caused by the pooling operation in Eq. \ref{eq:1}. Given the aforementioned attention map $M_{P}$ and the $Value$ matrix, the output GLocal feature can be expressed as:
\begin{equation}
F_{o} = \operatorname*{\textit{maxpooling}} \, (M_{P}V),
\end{equation}
where $V$ denotes the $Value$ matrix, i.e., $\mathcal{N}$. The detailed algorithm flow and feature dimension transformation of PPSA are shown in Fig. \ref{fig:PSA}. Lastly, we concatenate the $F_{o}$ at each scale to get the final point-wise GLocal feature $F_{P}$:
\begin{equation}
F_{P} = concat(F_{o1},F_{o2},F_{o3}),
\end{equation}
where $F_{o1},F_{o2},F_{o3}$ are generated from neighborhood feature maps at different scales.

\begin{figure*}[htbp]
  \centering
  \includegraphics[width=0.9\linewidth]{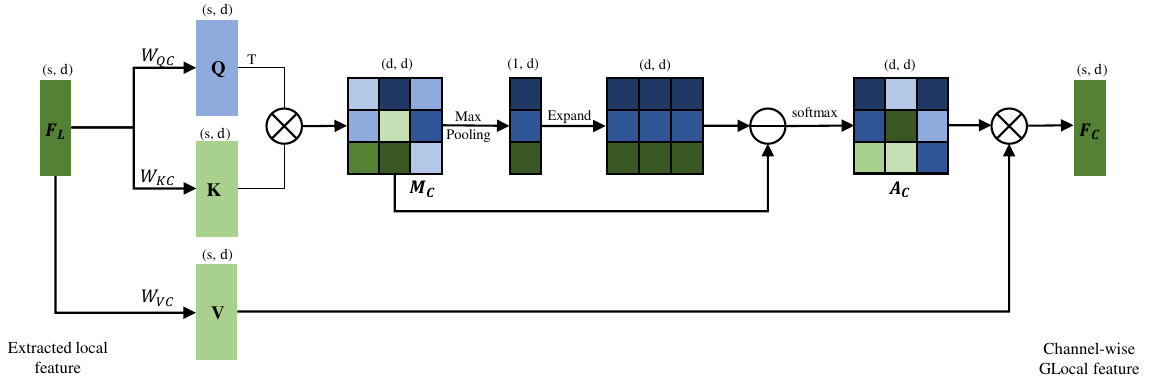}
  \caption{Channel-wise self-attention mechanism. An Affinity matrix $A_{C}$ is designed to avoid aggregating redundant features, enhancing the channel-wise GLocal feature representation.
  \label{fig:CSA}}
\end{figure*}

\textbf{Channel-wise Self-Attention.}
Apart from the PPSA mechanism, we also utilize the CSA mechanism to capture context dependencies in the channel dimension, enhancing the GLocal feature expression of our model. As shown in Fig. \ref{fig:CSA}, given the aggregated local feature $F_{L} \in R^{s\times d}$, we first compute the attention map $M_{C} \in R^{d \times d}$ of CSA as:
\begin{equation}
M_{C} = K^{T}Q = (F_{L}W_{KC})^{T}(F_{L}W_{QC}).
\end{equation}
where the shapes of $K, Q$ are reduced to $(s/8) \times d$ by weight matrices $W_{KC}$ and $W_{QC}$, to improve efficiency. Inspired by \cite{qiu2021geometric}, we calculate the affinity matrix $A_{C}$ based on $M_{C}$, to measure the difference among channels, which can be expressed as:
\begin{equation}
A_{C} = softmax(expand(maxpooling(M_{C}))-M_{C}),
\end{equation}
where $maxpooling(M_{C}) \in R^{d \times 1}$  extracts the maximum value of each row in $M_{C}$, $expand(\cdot)$ expands the matrix $maxpooling(M_{C})$ to the same size as $M_{C}$ by column repetition. From the subtraction, the larger in magnitude the element in $A_{C}$, the lower the similarity of the corresponding two channels. As such, CSA tends to focus on channels with significant differences, avoiding aggregating similar/redundant information. After that, we calculate the $Value$ matrix as:
\begin{equation}
V = F_{L}W_{VC},
\end{equation}
where $W_{VC}$ is a learnable weight matrix. Finally, the channel-wise GLocal feature $F_{C}$ can be expressed as:
\begin{equation}
F_{C} = VA_{C}.
\end{equation}

\begin{table*}[htbp]\color{black}
 \caption{Classification results on the ModelNet40 dataset \label{tab:cls_res}
 }
 \centering
 \begin{tabular}{c|c|c|c|c|c|c}
  \hline
   {Methods}  & Input Size & mAcc (\%) & OA (\%)  & {Parameters  (MB)} & {FLOPs (GB)} & {Frame Per Sec.}  \\
   \hline
  \multicolumn{7}{c}{Other Learning-based Methods} \\
  \hline
  {3DShapeNets}\cite{wu20153d} & 1024  & 77.3   & 84.7  & - & - & -     \\
  {PointNet}\cite{qi2017pointnet} & 1024  & 86.0   & 89.2     & 3.47 & 0.45 & \textbf{614} \\
  {PointNet++}\cite{qi2017pointnet++} & 1024  & 88.2   & 91.9    & 1.74 & 4.09 & 16     \\
  {diffConv}\cite{lin2021diffconv} & 1024 & 90.4   & 93.2   & 2.08 & \textbf{0.16} & -  \\
  {CurveNet}\cite{muzahid2020curvenet}  &1024  & 90.4   & 93.1  & - & - & -     \\
  {PointCNN}\cite{li2018pointcnn} & 1024  & 88.1   & 92.2   & \textbf{0.6} & 1.54 & 14 \\
  {DGCNN}\cite{wang2019dynamic} & 1024  & 90.2   & 92.2    & 1.81 & 2.43 & 279 \\
  {FatNet}\cite{kaul2021fatnet} & 1024  & 90.6   & 93.2    & - & - & -    \\
  {DRNet}\cite{qiu2021dense} & 1024  & -  & 93.1     & - & - & -    \\
  {PointNeXt}\cite{qian2022pointnext} & 1024  & 91.1  & 94.0 & 4.5 & 6.5 & -    \\
  {PointMLP}\cite{ma2022rethinking} & 1024  & 91.4  & \textbf{94.5} & 12.6 & - & 112    \\
  \hline
  \multicolumn{7}{c}{Transformer-based Methods} \\
  \hline
  {PATs}\cite{yang2019modeling} & 1024  & -  & 91.7    & - & - & -  \\
  {LFT-Net}\cite{gao2022lft} & 1024  & 89.7   & 93.2   & - & - & -   \\
  {PointTransformer}\cite{engel2021point} & 1024  & 89.0   & 92.8  & 13.86 & 9.36 & 17  \\
  {MLMST}\cite{han2021point} & 1024  & -  & 92.9   & - & - & -    \\
  {PointCloudTransformer}\cite{guo2021pct} & 1024  & 90.3   & 93.2   & \textbf{2.80} & \textbf{2.02} & \textbf{125}     \\
  {LSLPCT}\cite{song2022lslpct} & 1024  & 90.5   & 93.5   & - & - & -     \\
  {PointTransformer}\cite{zhao2021point}  & 1024  & 90.6   & 93.7    & 9.14 & 17.14 & 15   \\
  {CloudTransformers}\cite{mazur2021cloud} & 1024  & 90.8   & 93.1 & 22.91 & 12.69 & 12     \\
  {GBNet}\cite{qiu2021geometric} & 1024  & 91.0   & 93.8  & 8.38 & 9.02 & 102      \\  {3DCTN}\cite{lu20223dctn} & 1024  & 91.6   & 93.2  & 4.21 & 3.76 & 24      \\
  \hline
  {Ours}    & 1024  & \textbf{92.4 }  &\textbf{ 94.0}  & 5.21 & 3.09 & 15      \\

  \hline
 \end{tabular}
\end{table*}

Given both $F_{P}$ and $F_{C}$, the final GLocal feature can be generated by combining them with an element-wise addition:
\begin{equation}
F_{G} = F_{P} + F_{C}.
\end{equation}
Additionally, we apply a residual connection between the LFA block and the GFL block:
\begin{equation}
F_{OUT} = F_{L} + LBR(F_{G}),
\end{equation}
where $F_{OUT}$ is the final output feature map of the defined GFL block, and $LBR$ denotes the combination of $Linear$, $BatchNorm$, and $ReLU$ layers.

\section{Experiments}
\label{sec:Experiments}
In this section, we first introduce the implementation details of our 3DGTN, including hardware configuration and hyperparameter settings. 
Secondly, we present the performance evaluation of our method on the classification and segmentation tasks, comparing it to state-of-the-art methods.
Specifically, for the classification task, we tested our method on the widely-used ModelNet40 dataset \cite{wu20153d}. For the object part segmentation task, we tested our method on the ShapeNet dataset \cite{yi2016scalable}. For the semantic segmentation, we tested our method on the challenging Toronto-3D dataset \cite{tan2020toronto}.
Finally, we present the ablation experiment results on the main components of our method.

\subsection{Implementation Details}
We implemented our classification and segmentation networks in PyTorch. Both were trained and tested on an NVIDIA Tesla V100 GPU. We used the SGD Optimizer with a momentum of 0.9 and weight decay of 0.0001. The initial learning rate was set to 0.01, with a cosine annealing schedule to adjust the learning rate for each epoch. We trained classification, part segmentation, and semantic segmentation networks for 250, 300, and 500 epochs respectively, with the same batch size of 16.

\begin{figure*}[b]
  \centering
  \includegraphics[width=0.7\linewidth]{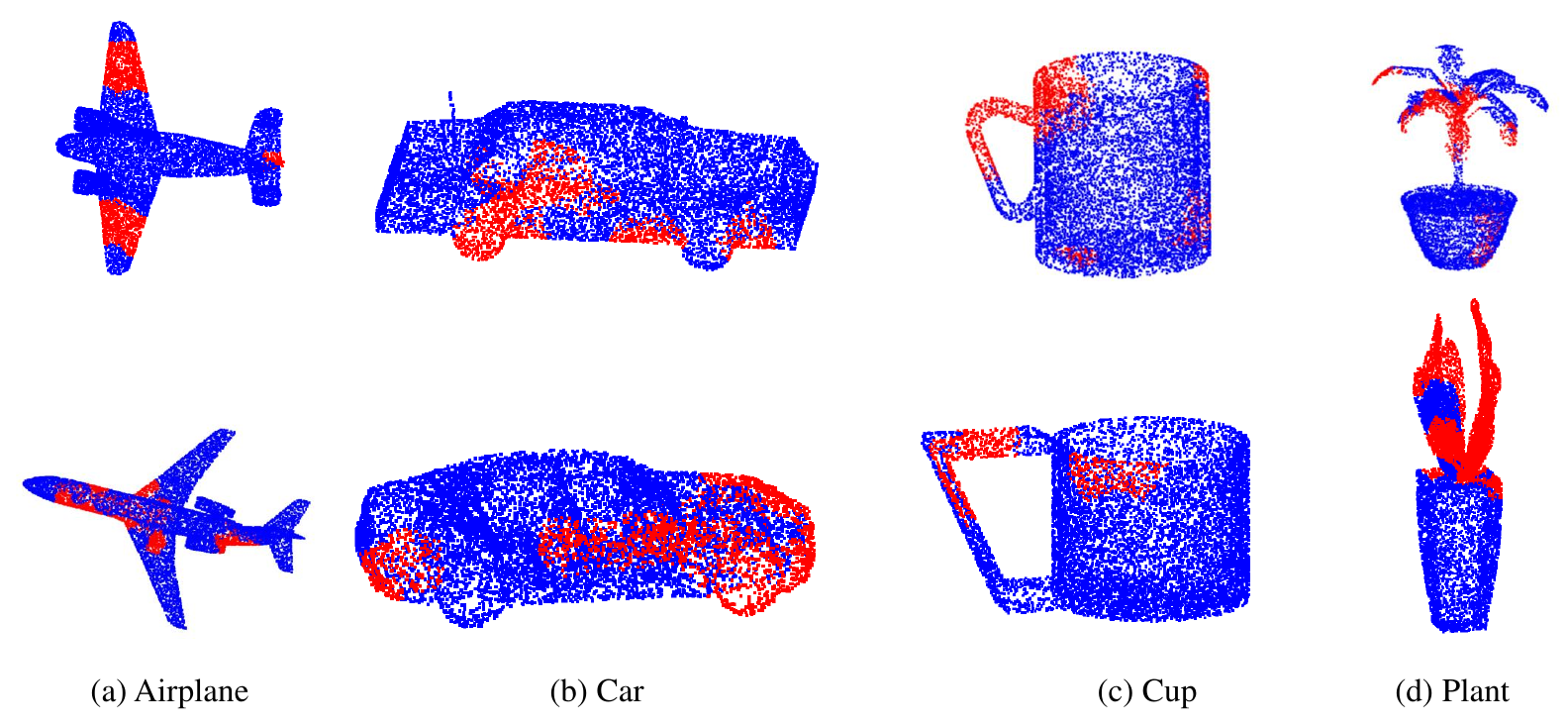}
  \caption{Heat map visualization of classification results. As can be seen, the attention (red) is focused on the discriminative parts of targets, such as the wings of an airplane, the tires of a car, the handle of a cup, and the leaves of a plant.
  \label{fig:heatmap}}
\end{figure*}

\subsection{Classification on ModelNet40 Dataset}
\textbf{Dataset and Metric.} 
The ModelNet40 dataset contains 12311 CAD models with 40 object categories. We split them into 9843 training samples and 2468 testing models, following PoineNet++ \cite{qi2017pointnet++}. For a fair comparison, we downsampled each input point cloud to 1024 points with normals by FPS. Since point clouds in ModelNet40 are generated from 3D meshes, we can easily obtain the normal of each point according to the corresponding face normal.
The mean accuracy within each category (mAcc) and the overall accuracy (OA) are used for performance evaluation.
Additionally, we adopt the total number of parameters, FLOPs (FLOating Point operations), and Frame Per Second to evaluate the model size and efficiency. 

\textbf{Performance Comparison.} 
We compared our 3DGTN with the state-of-the-art Transformer-based methods and other deep learning-based methods. The comparison results are shown in Table.  \ref{tab:cls_res}. As can be seen, our method achieves the best mean accuracy of 92.4\% among all benchmarked methods in terms of mAcc, outperforming the prior state-of-the-art (PointMLP \cite{ma2022rethinking}) by 1.0 absolute percentage points. In terms of OA, our method achieves the best result of 94.0\% among the Transformer-based methods. For the model size, our method requires fewer parameters (5.21MB) and FLOPs (3.09GB) compared to most Transformer-based algorithms, accounting for only 57\% and 18\% of Point Transformer \cite{zhao2021point} respectively. However, due to the naive implementation of several time-consuming operations like downsampling and $k$NN neighborhood construction, the inference speed of our method can still be improved.

\textbf{Heat Map Visualization.}
Fig. \ref{fig:heatmap} shows heat map visualization results to verify the interpretability of our method. Specifically, we obtain the regions of interest of our network for several point clouds of the Airplane, Car, Cup, and Plant classes by the Grad-CAM method \cite{zhou2016learning}. From the results, the attention (colored in red) is mainly focused on the wings and tail of the Airplane, the tires of the Car, the handle of the Cup, and the leaves of the Plant. As we can see, all the regions of interest are consistent with the human visual system, demonstrating our method's interpretability.

\begin{table}[htbp]\color{black}
 \caption{Part segmentation results (\%) on the ShapeNet dataset \label{tab:partseg_res}
 }
 \centering
 \begin{tabular}{c|c|c}
  \hline
   {Methods} & Cat. mIoU & Ins. mIoU  \\
  \hline
  {PointNet}\cite{qi2017pointnet} & 80.4  & 83.7    \\
  {A-SCN}\cite{xie2018attentional} & -  & 84.6    \\
  {PointNet++}\cite{qi2017pointnet++} & 81.9  & 85.1    \\
  {PCNN}\cite{atzmon2018point} & 81.8  & 85.1    \\
  {SpiderCNN}\cite{xu2018spidercnn} & 82.4  & 85.3    \\
  {SPLATNet}\cite{su2018splatnet} & 83.7  & 85.4   \\
  {Point2Sequence}\cite{liu2019point2sequence} & -  & 85.2   \\
  {DGCNN}\cite{wang2019dynamic} & 82.3  & 85.2    \\
  {SGPN}\cite{wang2018sgpn} & 82.8  & 85.8   \\
  {SubSparseCNN}\cite{graham20183d} & 83.3  & 86.0   \\  
  {PointCNN}\cite{li2018pointcnn} & 84.6  & 86.1   \\
  {PointConv}\cite{wu2019pointconv} & 82.8  & 85.7   \\
  {PVCNN}\cite{liu2019point} & -  & 86.2   \\
  {RS-CNN}\cite{liu2019relation} & 84.0  & 86.2   \\
  {KPConv}\cite{thomas2019kpconv} & 85.0  & 86.2    \\
  {InterpCNN}\cite{mao2019interpolated} & 84.0  & 86.3    \\
  {DensePoint}\cite{liu2019densepoint} & 84.2  & 86.4   \\
  {PAConv}\cite{xu2021paconv} & 84.6  & 86.1    \\
  {PointTransformer}\cite{zhao2021point} & 83.7  & \textbf{86.6}   \\
  {StratifiedTransformer}\cite{lai2022stratified} &\textbf{ 85.1}  & \textbf{86.6}    \\

  \hline
  {Ours}    & 84.0  & \textbf{86.6 }    \\

  \hline
 \end{tabular}
\end{table}

\subsection{Part Segmentation on ShapeNet Dataset}
\textbf{Dataset and Metric.} 
The ShapeNet dataset contains 16880 models with 16 shape categories. We split them into 14006 training samples and 2874 testing models, following Point Transformer \cite{zhao2021point}. The dataset has 50 part labels, and each object has at least two parts. For a fair comparison, we downsampled each input point cloud to 2048 points with normals by FPS. 
The category-wise mean Intersection over Union (mIoU) and instance-wise mIoU are used for performance evaluation.

\begin{figure*}[b]
  \centering
  \includegraphics[width=0.9\linewidth]{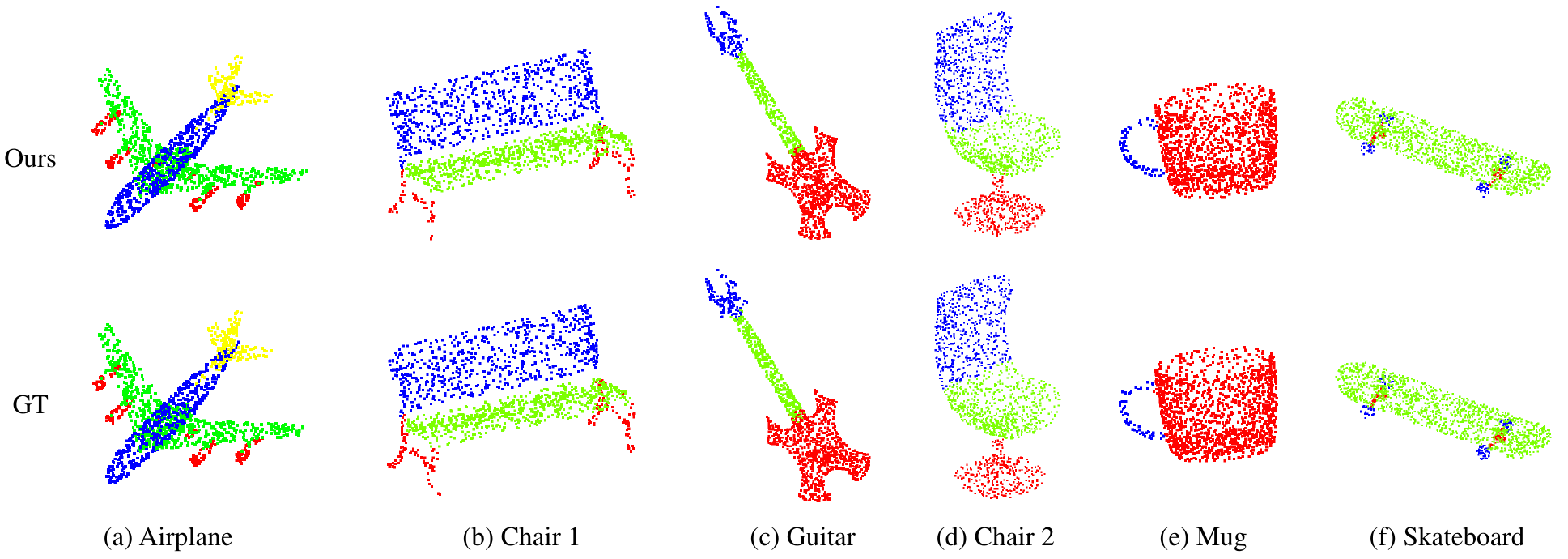}
  \caption{Part segmentation results from the ShapeNet dataset. As can be seen, our segmentation predictions are faithful to the ground truth.
  \label{fig:seg_visual}}
\end{figure*}

\textbf{Performance Comparison.} 
The comparison results are shown in Table. \ref{tab:partseg_res}. As measured by instance-wise mIoU, our 3DGTN achieves competitive results (86.6\%) compared with the SOTA Transformer-based methods such as Stratified Transformer \cite{lai2022stratified}. 
Several part segmentation results are shown in Fig. \ref{fig:seg_visual}.

\begin{table*}[htbp]\color{black}
 \caption{Semantic segmentation results (\%) on the Toronto-3D dataset \label{tab:semseg_res}
 }
 \centering
 \begin{tabular}{c|c|c|c|c|c|c|c|c|c}
  \hline
   {Methods}  & mIoU & Road & Road mrk.  & Natural & Building & Util.line & Pole & Car & Fence  \\
  \hline
  {PointNet++}\cite{qi2017pointnet++} & 56.55  & 91.44   & 7.59  & 89.80 &74.00 & 68.60 & 59.53 & 52.97 & 7.54     \\
  {PointNet++ (MSG)}\cite{qi2017pointnet} & 53.12  & 90.67   & 0.00  & 86.68 &75.78 & 56.20 & 60.89 & 44.51 & 10.19     \\
  {DGCNN}\cite{wang2019dynamic} & 49.60  & 90.63   & 0.44  & 81.25 &63.95 & 47.05 & 56.86 & 49.26 & 7.32     \\
  {KPFCNN}\cite{thomas2019kpconv} & 60.30  & 90.20   & 0.00  & 86.79 &\textbf{86.83} & 81.08 & 73.06 & 42.85 & 21.57     \\
  {MS-PCNN}\cite{ma2019multi} & 58.01  & 91.22   & 3.50  & 90.48 &77.30 & 62.30 & 68.54 & 52.63 & 17.12     \\
  {TG-Net}\cite{li2019tgnet} & 58.34  & 91.39   & 10.62  & 91.02 &76.93 & 68.27 & 66.25 & 54.10 & 8.16     \\
  {MS-TGNet}\cite{tan2020toronto} & 60.96  & 90.89   & 18.78  & \textbf{92.18} &80.62 & 69.36 & 71.22 & 51.05 & 13.59     \\
  {PointCloudTransformer}\cite{guo2021pct} & 79.32  & 79.77   & 59.51  & 75.78 &84.29 &77.78 &82.00 & 79.51 & 95.92     \\
  {diffConv}\cite{lin2021diffconv} & 76.73  & \textbf{83.31}   & 51.06 &69.04 &79.55 & 80.48 & 84.41 & 76.19 & 89.83     \\
  \hline
  {Ours}    & \textbf{82.53}  & 83.18  & \textbf{59.51} &84.70 &86.03 & \textbf{83.98} & \textbf{85.19} & \textbf{81.14} & \textbf{96.17}     \\

  \hline
 \end{tabular}
\end{table*}

\subsection{Semantic Segmentation on Toronto-3D Dataset}
\textbf{Dataset and Metric.} 
Toronto-3D is a challenging real-scene dataset, collected by a 32-line LiDAR sensor in large-scale urban outdoor scenarios. It covers approximately 1 km of road and consists of about 78.3 million points, labeled into 8 different categories. The dataset has been divided into four subsets: $L001$, $L002$, $L003$, and $L004$, where $L002$ was taken as the testing set, and the others were taken as the training set. We further divided each subset into $5m \times 5m$ blocks and sampled 2048 points with normals in each block, following diffConv \cite{lin2021diffconv}. The category-wise mIoU is used for performance evaluation. Additionally, the IoU for each category is also provided.

\textbf{Performance Comparison.} 
The comparison results are shown in Table. \ref{tab:semseg_res}. As can be seen, our 3DGTN outperforms all benchmarked methods in terms of category-wise mIoU. It surpasses the prior SOTA methods such as diffConv by 5.7 absolute percentage points. The Table also shows the specific IoU of each category, where our method achieves the best results in five of the eight categories, even for small objects like road marks. The results demonstrate that our method has excellent performance when processing real-scanned data, exceeding previous SOTA.  

\subsection{Ablation Study}
We conducted a series of ablation experiments for the main components of our 3DGTN to verify their effectiveness. These experiments were performed on the ModelNet40 dataset.

\begin{table*}[htbp]\color{black}
 \centering
  \renewcommand{\arraystretch}{1.5} 
 \caption{\textcolor{black}{Results of ablation study} 
 }
 \label{tab:ablation}
 \begin{tabular}{c|c|c|c|c|c|c}
  \hline
    \multicolumn{2}{c|}{Ablation}  & mAcc (\%)  & OA (\%) & Parameters (MB)  & FLOPs (GB) & Frame Per Sec. \\
 \hline

  \multirow{2}{*}{Local feature aggregation} & Graph convolution $\rightarrow$ Standard MLP & 91.6  & 93.1 & 5.18 & 0.45 & 17     \\
  \cline{2-7}
      &Multi-scale $\rightarrow$ Single-scale &91.3  & 92.9 & 2.11 & 0.64 & 19 \\
   \hline
   \multirow{4}{*}{GLocal feature learning} & $-$ & 91.2  & 92.7 & 2.25 & 2.78 & 17    \\
  \cline{2-7}
      &$-$ CSA & 91.9  & 93.7 & 4.71 & 3.04 & 16 \\
  \cline{2-7}
     &$-$ PPSA & 91.3  & 93.2 & 3.75 & 2.95 & 16 \\
  \cline{2-7}
    &PPSA $\rightarrow$ Vanilla PSA &91.5  & 93.6 & 5.11 & 3.08 & 15 \\
    \hline
  {Multi-level GLocal feature} & $-$ & 92.1  & 93.6 & 4.75 & 3.09 & 15     \\
   \hline
   \hline

  \multicolumn{2}{c|}{3DGTN}  & \textbf{92.4}  & \textbf{94.0} & 5.21 & 3.09 & 15  \\
  \hline
 \end{tabular}
\end{table*}



\textbf{Local Feature Aggregation Block.} 
We first investigate the effectiveness of the LFA block, which is used to capture local information. As shown in Table. \ref{tab:ablation} Row 2, the performance with the MLP-based LFA block is 91.6\%/93.1\% in terms of mAcc/OA, which is lower than that with the initial LFA block (92.4\%/94.0\%). This demonstrates that the GCN-based LFA block plays an important role in our algorithm. We also replaced the multi-scale strategy of the LFA block with the single-scale one. From the result in Table. \ref{tab:ablation} Row 3, the classification performance of the multi-scale strategy is better than the single-scale strategy (91.3\%/92.9\%). This suggests that the multi-scale features are beneficial to enhancing the expression of local information, thereby improving the performance of our algorithm.

\textbf{GLocal Feature Learning Block.} 
We conducted a detailed ablation study on the GFL block. As shown in Table. \ref{tab:ablation}, firstly, we removed the GFL block. As can be seen, the performance drops significantly, which demonstrates that the GFL block is essential to our algorithm. Secondly, since the GFL block contains two important mechanisms: PPSA and CSA, we also studied the effectiveness of each mechanism. When the CSA was removed, the classification accuracy (mAcc/OA) drops from 92.4\%/94.0\% to 91.9\%/93.7\%. Likewise, when the PPSA was removed, there is a similar drop (from 92.4\%/94.0\% to 91.13\%/93.2\%). These results suggest that both self-attention mechanisms are effective in improving classification performance. Additionally, to further verify the effectiveness of the PPSA mechanism, we replaced it with a regular point-wise self-attention mechanism (treating the $F_{L}$ as the $Value$ matrix). After replacing, we observe a 0.9\% and 0.4\% drop in mAcc and OA respectively. This confirms the superiority of our PPSA mechanism.

\textbf{Multi-level GLocal Feature Concatenation.}
We studied the effectiveness of the multi-level GLocal feature. As illustrated in Fig. \ref{fig:overview}, we concatenate the output GLocal feature of each level (module) by a residual connection to generate the multi-level GLocal feature. As shown in Table. \ref{tab:ablation} Row 8, when the residual connection was removed, we observe a 0.3\% and 0.4\% drop in mAcc and OA respectively. This suggests that the multi-level GLocal feature contributes significantly to performance improvement.

\section{Conclusion}
\label{sec:conclusion}
In this paper, we have proposed a hierarchical point cloud representation network for classification and segmentation, named 3DGTN.
Taking the point cloud, either with normals as input, the encoder first projects it into a higher-dimensional space by using stem MLP, followed by feeding the projected feature map into several cascaded LFA and GFL blocks for GLocal feature extraction. The LFA block is achieved by using multi-scale GCN. For the GFL block, we adopt the dual-attention Transformer for the GLocal feature expression, which combines the PPSA and CSA mechanisms. Specifically, the novel PPSA mechanism is designed to fuse both global features and local neighborhood information of input points, which is able to improve GLocal feature learning ability and mitigate local information loss. Then, the GLocal features from the GFL blocks at different modules are concatenated, followed by being fed into an additional Graph Convolution layer and further a Max-pooling layer to generate the final classification results. Additionally, taking the generated GLocal feature as input, we design an All-MLP decoder for the segmentation task.
Extensive experiments on the ModelNet40 classification dataset \cite{wu20153d}, ShapeNet part segmentation dataset \cite{yi2016scalable}, and Toronto-3D semantic segmentation dataset \cite{tan2020toronto} demonstrate the superiority of our method in dealing with both synthetic and real-scene data. 

\textbf{Future Work.} Our hierarchical network uses Euclidean distance-based downsampling and neighborhood search methods, which are time-consuming and cannot serve the semantic information extracted by the network very well. Since the attention map in the Transformer contains rich feature relationships, we plan to utilize the attention map for semantic-based point cloud sampling and grouping as a future research project. To this end, the ``superpoint" strategy could be a potential solution. 

\bibliographystyle{IEEEtran}
\bibliography{mybibfile}

\vfill

\end{document}